\documentclass[conference]{IEEEtran}
\IEEEoverridecommandlockouts
\usepackage{cite}
\usepackage{amsmath,amssymb,amsfonts}
\usepackage{algorithmic}
\usepackage{graphicx}
\usepackage{textcomp}
\usepackage{xcolor}

\usepackage{physics}
\usepackage{caption}
\usepackage{url}
\usepackage{balance}
\usepackage{hyperref}
\usepackage{multirow}

\renewcommand{\L}{\mathcal{L}}

\graphicspath{{./figures/}}

\def\BibTeX{{\rm B\kern-.05em{\sc i\kern-.025em b}\kern-.08em
		T\kern-.1667em\lower.7ex\hbox{E}\kern-.125emX}}
\begin{document}
	
	\title{Meshless method stencil evaluation with machine learning}
	
	\author{
		\IEEEauthorblockN{Miha 
		Rot\IEEEauthorrefmark{1}\IEEEauthorrefmark{2}, Aleksandra 
		Rashkovska\IEEEauthorrefmark{2}}
		
		\IEEEauthorblockA{\IEEEauthorrefmark{1}Jozef Stefan International 
			Postgraduate School, Jamova cesta 39, 1000 Ljubljana, Slovenia}
		\IEEEauthorblockA{\IEEEauthorrefmark{2}``Jožef Stefan'' Institute, 
			Parallel and Distributed Systems Laboratory, Jamova cesta 39, 1000 
			Ljubljana, Slovenia\\
			miha.rot@ijs.si}
		
		\thanks{The authors would like to acknowledge the financial
			support of the Slovenian Research Agency (ARRS) research core 
			funding No. P2-0095, research project J2-3048, and the Young Researcher program PR-10468.}
	}
	
	\maketitle

\begin{abstract}
	Meshless methods are an active and modern branch of numerical analysis 	
	with many intriguing benefits. One of the main open research questions 	
	related to local meshless methods is how to select the best possible 	
	stencil - a collection of neighbouring nodes - to base the calculation on. 	
	In this paper, we describe the procedure for generating a labelled stencil 
	dataset and use a variation of pointNet - a deep learning network based on 
	point clouds - to create a classifier for the quality of the stencil. We 
	exploit features of pointNet to implement a model that can be used to 
	classify differently sized stencils and compare it against models dedicated 
	to a single stencil size. The model is particularly good at detecting the 
	best and the worst stencils with a respectable area under the curve (AUC) 
	metric of around 0.90. There is much potential for further improvement and 
	direct application in the meshless domain.
\end{abstract}

\bigskip



\section{Introduction}
Partial differential equations (PDE) are one of the most common approaches 
to modelling natural phenomenon and industrial processes. Closed-form 
solutions are rare, leaving us to rely on numerical approaches. Meshless 
methods are a relatively novel approach to PDE solving. They ease domain 
discretisation by avoiding meshing problems that arise in higher dimensions 
when using the traditional finite elements and finite volume 
methods~\cite{liu2010intro}. The Radial Basis Function Generated Finite 
Differences (RBF-FD)\cite{tolstykh2003using} method is used to calculate an 
approximation of a linear operator $\L$ applied to field $u$. 
Discretisation is achieved by populating the domain with computational 
nodes that hold field values $u_i$. The operator value in $i$-th node is 
calculated based on field values in $s$ neighbouring nodes that form its 
computational stencil $S_i$ 
\begin{equation}
	(\L u)_i \approx \sum_{j=1}^{s} w_{i, j} u_{S_i(j)},
	\label{eq:operatorApprox}
\end{equation}
where $w_{i, j}$ are the precomputed approximation coefficients. The 
coefficients are determined by demanding the equation~\eqref{eq:operatorApprox} 
to be exact for a set of basis functions and solving the resulting  system.

One of the factors contributing to the approximation accuracy are the positions 
of nodes in the computational stencil. Due to the many confounding factors it 
is difficult to predict the stencil's quality before solving the weight system 
and evaluating the operator. Stencil nodes are usually selected as the $s$ 
closest nodes to the one we are approximating the operator in. This often 
proves to be inadequate, especially when close to the domain boundary or 
dealing with varying node density. Existing attempts that tackle the problem 
use, amongst other things, linear programming~\cite{Seibold2008minimalPositive} 
and geometric assumptions~\cite{davydov2022improved}.

In this paper, we use a machine learning approach to devise a method that can 
estimate the approximation quality of the stencil without solving the 
weight system. Such method, if implemented efficiently, can be used to 
construct better stencils and evaluate the quality of discretisation before 
proceeding to the subsequent computationally demanding step of approximation 
building.

The rest of the paper is organised as follows. In Section~\ref{ch:data} we 
describe the dataset and explain how it was constructed, followed by the 
description of the utilised machine learning method in Section~\ref{ch:method}. 
Next, we present the results in Section~\ref{ch:results}. We conclude the paper 
with directions for future work in Section~\ref{ch:future} and the conclusion 
in Section~\ref{ch:conclusions}.

\section{Data preparation}
\label{ch:data}

The dataset is created algorithmically by generating a set of random 
stencils that are still representative of the stencils that could appear 
during a normal solution procedure. The RBF-FD algorithm, using the polyhamonic 
$r^3$ radial basis function (RBF) with $2$-nd order monomial 
augmentation\cite{jancic2021monomial}, is implemented in Medusa~\cite{medusa} 
C++ library for meshless PDE solving and used to approximate the gradient and 
the Laplace operator applied to a selection of fields with known analytic 
derivatives. The sum of absolute offsets between the approximated and the exact 
operator values for both operators on all test fields is then used as the measure of error $\epsilon$ that is used as a label for the stencil's fitness, with a lower number signifying a better node 
configuration.

\begin{figure}
	\includegraphics[width=\linewidth]{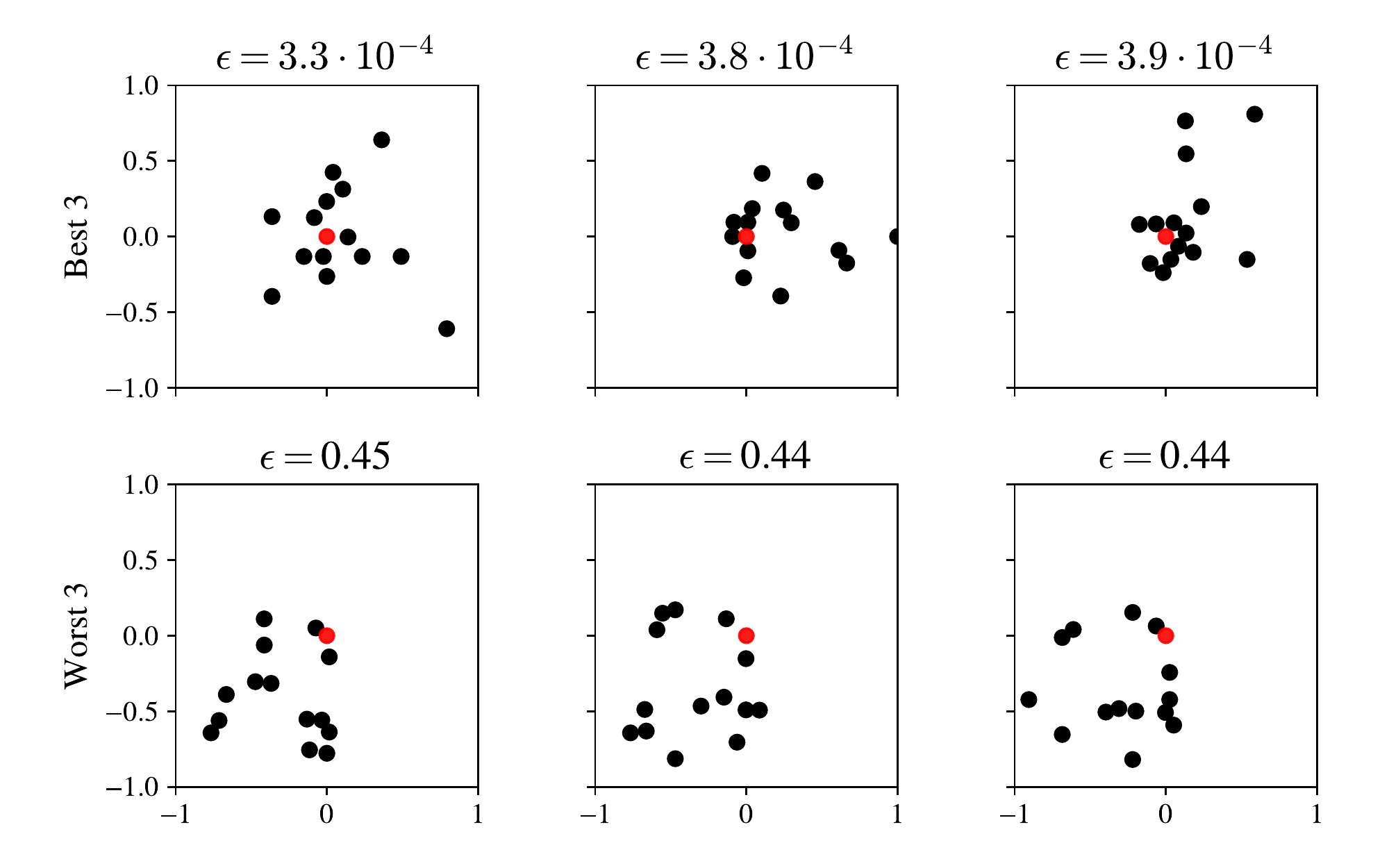}
	\caption{The best 3 stencils with $s = 15$ in the top row and the worst 
		3 in the bottom. The red point marks the central node where we 
		approximate the operator. Positions are centred and normalised.}
	\label{fig:bestWorstStencils15}
\end{figure}

Examples of the best and the worst stencils with size $s=15$ are shown in 
Fig.~\ref{fig:bestWorstStencils15}. Good stencils are relatively centred 
and symmetric as expected, while the bad ones exhibit lines of nodes that 
are unable to provide a good description of the underlying field.

We generate $10^5$ stencils and corresponding labels for sizes of $s \in \{6, 
7, 9, 
12, 15\}$ and combine them into a mixed size dataset that will be denoted as 
"mix". Then, we 
generate additional $5 \cdot 10^5$ stencils for sizes of $s \in \{9, 15\}$ as 
single-sized stencil datasets. The automatically generated dataset allows us to 
extend the size of the dataset in 
future work as far as required within the computational and time constraints. Creating larger 
datasets might be beneficial because choosing even the best among $5 \cdot 
10^5$ nodes, as shown in 
Fig.~\ref{fig:bestWorstStencils15}, does not yield even close to perfect 
stencils.

\begin{figure*}
	\includegraphics[width=\linewidth]{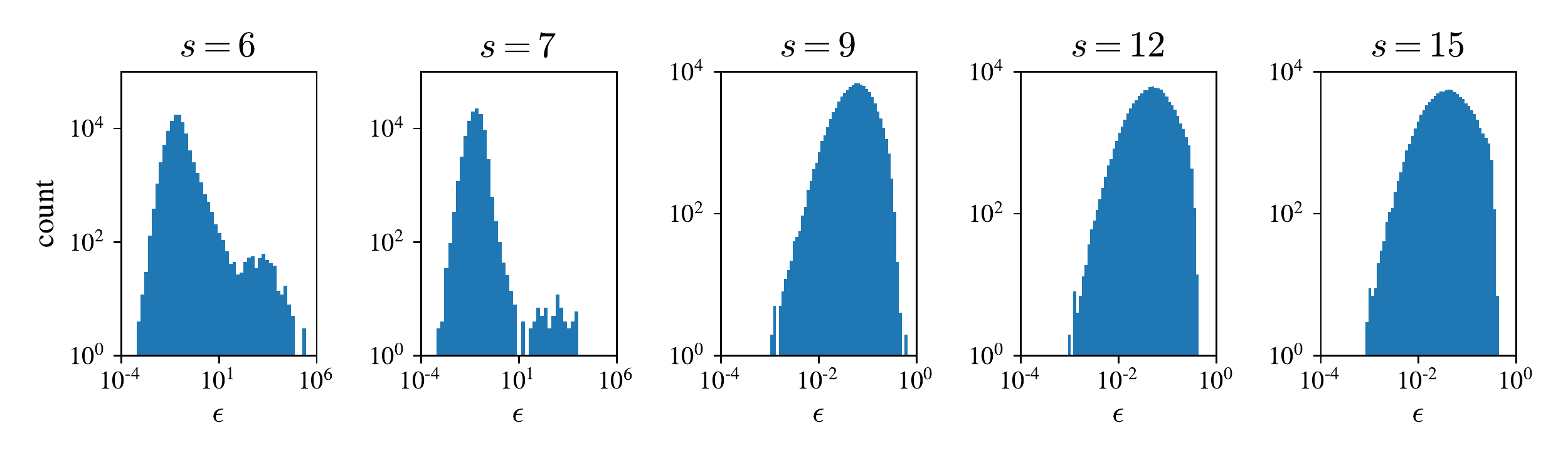}
	\caption{Distribution of error measure $\epsilon$ in datasets for 
		stencils with different sizes $s$. The smaller stencils have a much 
		higher variation.}
	\label{fig:errorHistograms}
\end{figure*}

The distributions of resulting $\epsilon$ labels for different stencil size 
datasets are 
shown in Fig.~\ref{fig:errorHistograms}. The distributions on the left 
figures for $s = 6$ and $s = 7$ provide an additional motivation for this 
endeavour. Small stencil size is preferred for meshless approximation due to the inherently 
lower computational costs that are a function of $s$ at every operator 
evaluation. As this 
graph shows, small stencils exhibit a much higher variance in quality when compared to their 
larger counterparts. That makes them less stable and harder to construct when not aided by 
advanced algorithms.

We discretise the continuous error label into 4 quartiles with equal number of 
stencils that guarantee a balanced 
dataset. Some care needs to be taken when determining classes for the "mix" dataset to 
maintain consistency. Quartile borders have to be created separately for each of the 
constituent stencil sizes to account for the different ranges of error labels seen in 
Fig.~\ref{fig:errorHistograms}.

\subsection{Stencil generation}
Random stencil selection needs to be designed in a way that replicates the 
stencils that are likely to appear in actual use. Purely random node 
placement is unsuitable due to the possibility of extremely close nodes that 
would cause instabilities for the approximation. This is established 
knowledge in the field~\cite{slak2019generation} and the availability of fast 
and efficient algorithms for generation of candidate nodes prevents this from 
occurring in practice. On the other hand, using available algorithms for node 
placement leads to stencils that are too uniform to provide any insight for 
machine learning.

We circumvent this problem by using an established meshless node 
positioning algorithm~\cite{slak2019generation} to generate a dense 
discretization of candidate nodes. The central node is randomly selected from 
the candidates. The rest of the stencil is then randomly sampled from the 
candidate nodes with a slightly radially decreasing probability to 
encourage some extent of aggregation. All $s$ nodes from this stencil are 
then alternately used as centres when constructing the approximation to 
introduce more non-symmetric stencils into the mix. We are specially 
interested in non-symmetric stencils as those are commonly found close to 
the domain boundaries where stencil construction is the most problematic.

This process is repeated until a desired number of stencils is generated. 
Stencil coordinates are then transformed into a coordinate representation with origin in the central node and normalized with the distance between the central and the farthest node.

\subsection{Test fields}
As mentioned in the beginning of this section, analytically differentiable 
fields are required as a benchmark for the RBF-FD approximation. The fields 
need to be diverse enough to provide a satisfactory proxy for fields 
encountered during the normal PDE solving and still simple enough to 
implement a general derivative in C++. We use a benchmark that consists of 
three fields:
\begin{itemize}
	\item \textbf{Monomial}
	\begin{equation}
		f(x, y) = x^ny^m, \qquad n=2, m=3;
	\end{equation}
	\item \textbf{Sinusoidal}
	\begin{equation}
		f(x, y) = \sin(k_x x) \sin(k_y y), \qquad k_x=2, k_y=1;
	\end{equation}
	\item \textbf{Exponential}
	\begin{equation}
		f(x, y) = e^{-\frac{x^2 + y^2}{2 \sigma}} , \qquad \sigma=1.
	\end{equation}
\end{itemize}

\section{Method}
\label{ch:method}

\begin{figure*}
	\includegraphics[width=\linewidth]{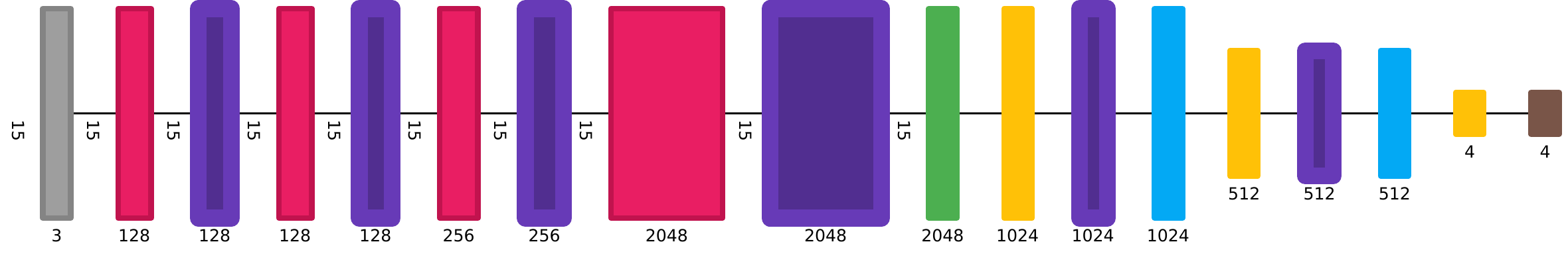}
	\includegraphics[width=\linewidth]{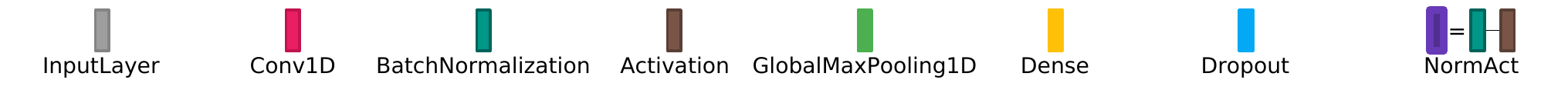}
	\caption{Modified PointNet neural network architecture for $s=15$. Visualisation created 
		directly from the Keras model using the Net2Vis~\cite{Net2Vis} framework.}
	\label{fig:networkGraph}
\end{figure*}

PointNet\cite{pointNet} was a groundbreaking deep learning algorithm that has opened the field for working with point cloud data. It is mainly used in 3D for classification and segmentation of LiDAR data. The point-based representation offers a much better storage efficiency compared to previously used methods such as voxel grids. The architecture of PointNet is relatively simple. The initial transformations of data lead into a single symmetric max-pooling layer, which is responsible for most of the many beneficial properties that led us to choose this model. In our	case, the permutation invariance and the support for internodal interactions were the most important properties.

Initially, we modified the final dense layers of PointNet to work for the task 
of regression, but the results were underwhelming and comparable to multilayer perceptrons. We settled on discretising our dataset and proceeding with a variation of the standard classification pointNet. The beneficial feature of the max-pooling layer to ignore the contribution of duplicate input coordinates, allowed us to pad the stencil dataset to the same size and utilise the same network for different stencil sizes. 

We used a slightly modified version of PointNet with the architecture shown in Fig.~\ref{fig:networkGraph}. The network consists of one-dimensional convolutional layers (Conv1D) with a kernel size of 1 and an increasing number of filters to increase the dimensionality of the input data. The convolutional layers are followed by a max-pooling layer (GlobalMaxPooling1D) that aggregates information from all stencil nodes and constructs a global feature vector used as an input for the dense (Dense) layers. The dense layers utilise dropout (Dropout) layers that randomly drop part of the coefficients and help against over-fitting. The convolutional and the dense layers are augmented with batch normalization and activation layers (NormAct). This version, named vanilla PointNet in the original paper, omits the two transformational layers in the steps before max-pooling. These layers were superfluous based on our testing and only added to the complexity without providing much benefit. This is most likely the case because the stencil data is already translated and well aligned. Instead, we doubled the size of all the other layers. The result is a similar number of weights as in the original network with transformation layers, but with better results in our use-case.

We implemented the modified PointNet in TensorFlow using the Adam optimiser and the sparse categorical crossentrophy loss function. We use ReLu on the intermediate activation layers and softmax on the output. A 0.3 dropout is used with the dense layers. The model was trained with a batch size of 1024 for 20 epochs. The stencil data was split into training and test partitions with 20\% intended for testing. 

\section{Results}
\label{ch:results}

The results section focuses on the comparison between the unified 
mixed stencil size model, implemented by padding smaller stencils, as mentioned 
in 
the previous chapter, and three single-size models. All models were 
trained with identically sized datasets with $5 \cdot 10^5$ stencils.

\begin{figure}
	\includegraphics[width=\linewidth]{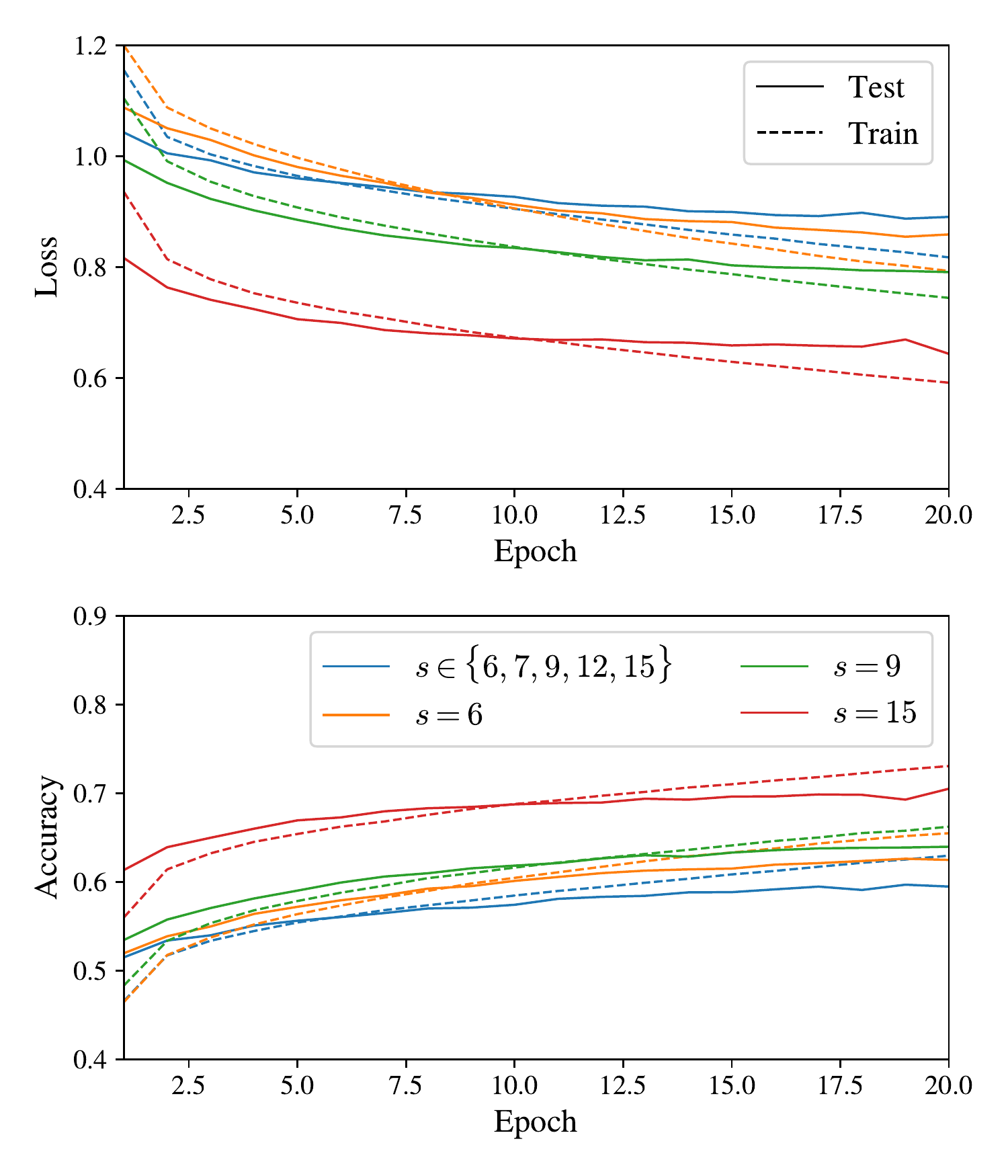}
	\caption{Decrease in loss function and increase in accuracy during the 
		training for three individual stencil sizes and the mix.}
	\label{fig:lossAccuracy}
\end{figure}

First, the training loss and accuracy curves are compared in 
Fig.~\ref{fig:lossAccuracy}. 
The unified model does have slightly worse results, but not drastically so. The accuracy for 
individual models increases with stencil size. Training is stopped at the 
arbitrarily selected 20th epoch. A fixed end 
point is used instead of a stopping criterion based on a validation dataset to ensure a direct 
comparison between the different models.

\begin{figure}
	\includegraphics[width=\linewidth]{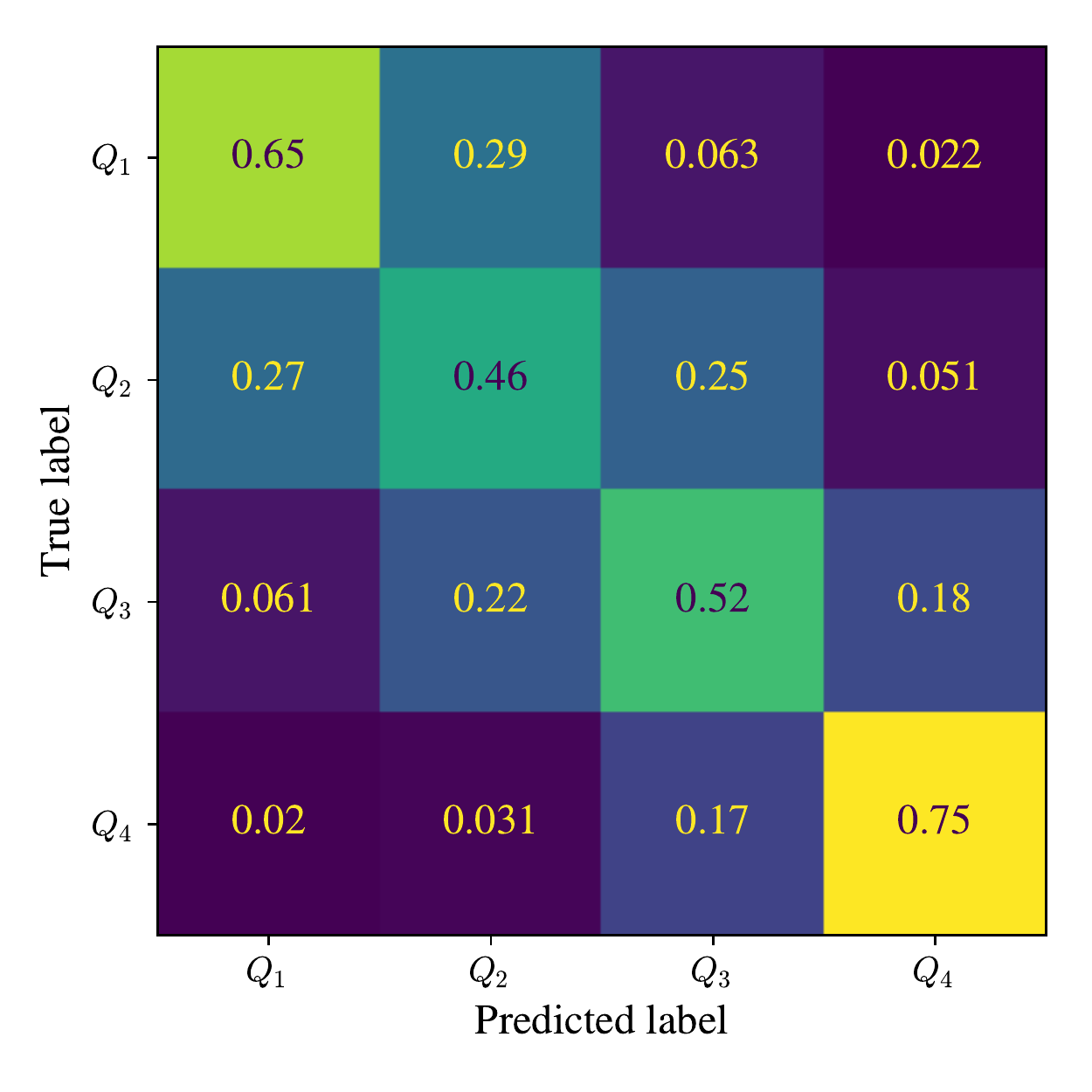}
	\caption{Confusion matrix with normalized columns for the mixed stencil size model with 
	$s \in \{6, 7, 9, 12, 15\}$. Values on the diagonal are the correctly identified classes.}
	\label{fig:mixConfusionMatrix}
\end{figure}

We take a closer look at the unified model and its confusion matrix shown in 
Fig.~\ref{fig:mixConfusionMatrix}. The columns show what true classes were 
correctly or incorrectly attributed to the given predicted class. The following discussion shows that the model performs much better than the 60\% accuracy would initially suggest. When analysing stencils, we are most interested in the best $Q_1$ class of stencils that we definitely want to use and the worst $Q_4$ class that we want to avoid at all cost. The classification accuracy is significantly higher for $Q_1$ and especially for $Q_4$ stencils, which are actually the two classes we are interested in. Additionally, most of the 
misclassifications for the predicted $Q_1$ were actually true $Q_2$ and thus 
still better than the median stencil and vice versa for $Q_4$. By definition of 
equal count quartiles, the stencil with median error falls onto the border 
between $Q_2$ and $Q_3$. Its error $\epsilon$ is still relatively small due to 
the shape of the distributions shown in Figure~\ref{fig:errorHistograms}. When 
accounting for this kind of a misclassification, 92\% of stencils classified as 
$Q_1$ were better, and 93\% of those classified as $Q_4$ were worse than the 
median stencil.

\begin{figure}
	\includegraphics[width=\linewidth]{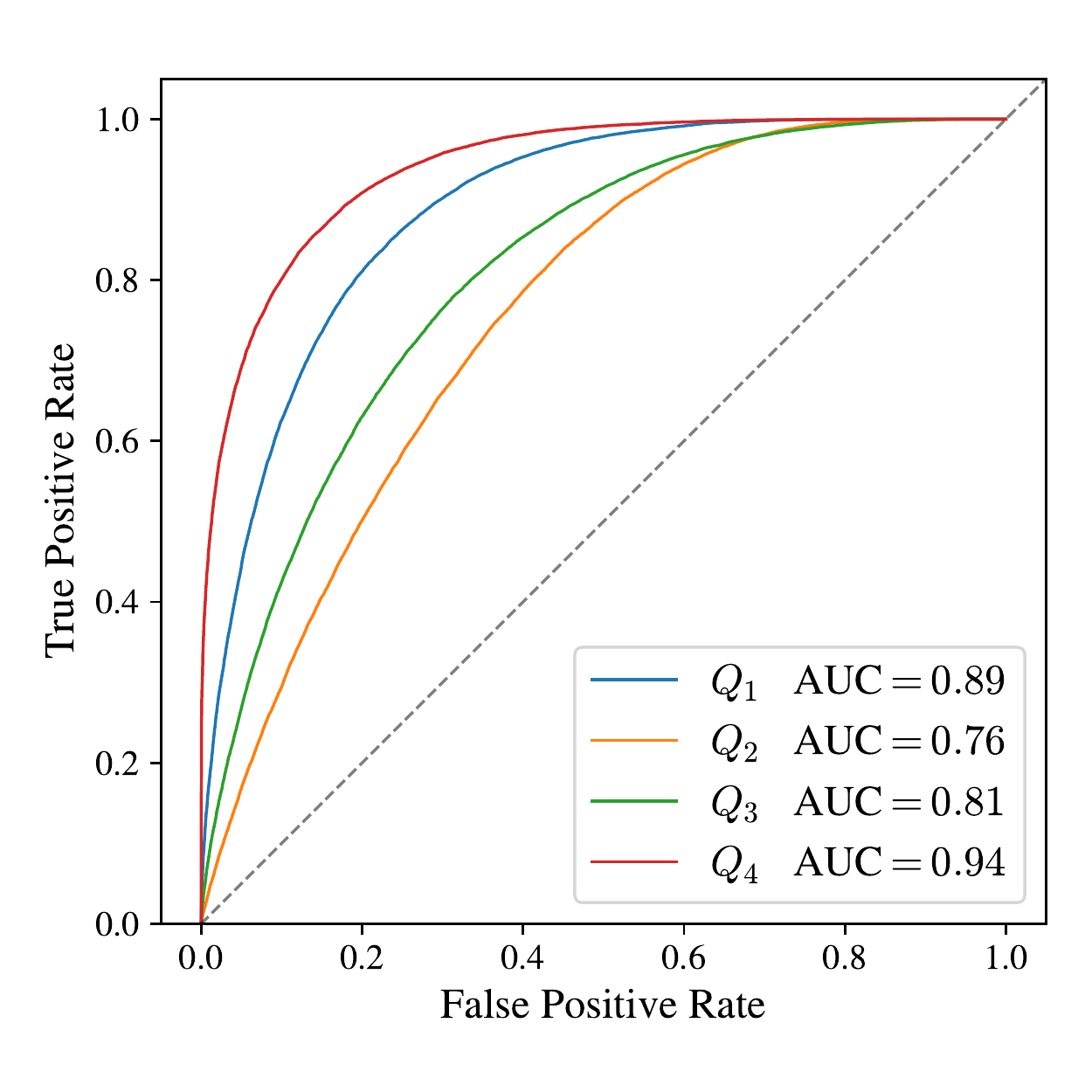}
	\caption{Receiver operating characteristic (ROC) curve for the 
		mixed stencil size model, with the area under the curve (AUC) as one of 
		the 
		classification quality metrics.}
	\label{fig:mixROC}
\end{figure}

We also look at the receiver operating characteristic (ROC) curve in 
Fig.~\ref{fig:mixROC}. 
The model does not return a class directly but a probability that the stencil belongs to any of the classes. The ROC 
curve plots the true positive rate (TPR) and the false positive rate (FPR) as the threshold 
probability for classification is varied. A perfect model would be a step function jumping to 
1 TPR as soon as the threshold becomes non-zero. We compare how close our 
models come to this 
ideal with a metric called area under the curve (AUC)~\cite{bradley1997auc} 
that confirms that our model is 
particularly good at detecting the bad $Q_4$ stencils with a respectable AUC 
metric of 0.94, and the good $Q_1$ stencils with a respectable AUC metric of 
0.89.

The main comparison between mixed and single stencil size models is presented in 
Table~\ref{tab:tabela} that holds the extended classification metrics for different 
variations of test and train datasets. We use four metrics:
\begin{itemize}
	\item \textbf{Accuracy} is the overall fraction of predictions that were 
	correct;
	\item \textbf{Precision} is the fraction of the predicted class that was correctly 
	classified;
	\item \textbf{Recall} is the fraction of the true class that was correctly classified;
	\item \textbf{F1} is the harmonic mean of precision and recall;
\end{itemize}
We notice two trends: (1) the classification results are significantly better for larger stencil 
sizes; (2) the mixed size model is consistently worse than the dedicated ones, with 
discrepancy increasing for smaller stencil size. The effect of increasing stencil size is 
stronger, with $s=15$ classification on the mixed model achieving better overall results than 
either $s=6$ or $s=9$ on the dedicated models.

The extreme classes remain favoured in classification accuracy but there is a reversal for 
the small $s = 6$ stencil. The best $Q_1$ is the most accurately classified class for this 
dataset achieving the best $Q_1$ precision among all the test cases. This could 
again be 
explained by the large range of error labels in Fig.~\ref{fig:errorHistograms} 
making the 
best quartile more distinct. This is encouraging because the small stencils, as mentioned 
previously, offer the biggest promise for direct application of this methodology.

The mixed size model provides consistently worse results than the dedicated 
models and 
involves additional frivolous computation when padding is required. But still, even with all 
of the aforementioned caveats, we consider it to be beneficial as it simplifies 
the stencil 
building procedure by streamlining the comparison of quality with and without a specific node.

\begin{table}
	\caption{Comparison of metrics between the single and multi-size training}
	\begin{tabular}{|ccc|cccc|}
		\hline
		Test & Train & Quartile & Precision & Recall & $F1$ & Accuracy \\
		\hline
		\multirow{4}{*}{$\text{mix}$} & \multirow{4}{*}{$\text{mix}$}
		& $Q_1$ & $0.65$    & $0.66$ & $0.65$ & \multirow{4}{*}{$0.60$} \\
		&& $Q_2$ & $0.46$    & $0.41$ & $0.43$ &                         \\
		&& $Q_3$ & $0.52$    & $0.55$ & $0.54$ &                         \\
		&& $Q_4$ & $0.75$    & $0.78$ & $0.76$ &                         \\
		\hline
		\multirow{4}{*}{$6$} & \multirow{4}{*}{$\text{mix}$}
		& $Q_1$ & $0.70$    & $0.64$ & $0.67$ & \multirow{4}{*}{$0.56$} \\
		&& $Q_2$ & $0.47$    & $0.46$ & $0.46$ &                         \\
		&& $Q_3$ & $0.50$    & $0.41$ & $0.45$ &                         \\
		&& $Q_4$ & $0.56$    & $0.71$ & $0.63$ &                         \\
		\hline
		\multirow{4}{*}{$6$} & \multirow{4}{*}{$6$}
		& $Q_1$ & $0.76$    & $0.72$ & $0.74$ & \multirow{4}{*}{$0.64$} \\
		&& $Q_2$ & $0.55$    & $0.52$ & $0.53$ &                         \\
		&& $Q_3$ & $0.57$    & $0.55$ & $0.56$ &                         \\
		&& $Q_4$ & $0.66$    & $0.77$ & $0.71$ &                         \\
		\hline
		\multirow{4}{*}{$9$} & \multirow{4}{*}{$\text{mix}$}
		& $Q_1$ & $0.66$ & $0.67$ & $0.67$ & \multirow{4}{*}{$0.61$} \\
		&& $Q_2$ & $0.49$    & $0.41$ & $0.44$ &                         \\
		&& $Q_3$ & $0.52$    & $0.57$ & $0.54$ &                         \\
		&& $Q_4$ & $0.77$    & $0.79$ & $0.79$ &                         \\
		\hline
		\multirow{4}{*}{$9$} & \multirow{4}{*}{$9$}
		& $Q_1$ & $0.66$ & $0.76$ & $0.71$ & \multirow{4}{*}{$0.64$} \\
		&& $Q_2$ & $0.51$    & $0.42$ & $0.46$ &                         \\
		&& $Q_3$ & $0.55$    & $0.57$ & $0.56$ &                         \\
		&& $Q_4$ & $0.82$    & $0.81$ & $0.81$ &                         \\
		\hline
		\multirow{4}{*}{$15$} & \multirow{4}{*}{$\text{mix}$}
		& $Q_1$ & $0.63$ & $0.74$ & $0.68$ & \multirow{4}{*}{$0.67$} \\
		&& $Q_2$ & $0.50$    & $0.40$ & $0.44$ &                         \\
		&& $Q_3$ & $0.63$    & $0.69$ & $0.66$ &                         \\
		&& $Q_4$ & $0.91$    & $0.85$ & $0.88$ &                         \\
		\hline
		\multirow{4}{*}{$15$} & \multirow{4}{*}{$15$}
		& $Q_1$ & $0.67$ & $0.77$ & $0.72$ & \multirow{4}{*}{$0.70$} \\
		&& $Q_2$ & $0.54$    & $0.51$ & $0.52$ &                         \\
		&& $Q_3$ & $0.68$    & $0.66$ & $0.67$ &                         \\
		&& $Q_4$ & $0.92$    & $0.85$ & $0.88$ &                         \\
		\hline
	\end{tabular}
	\label{tab:tabela}
\end{table}

\begin{figure}
	\includegraphics[width=\linewidth]{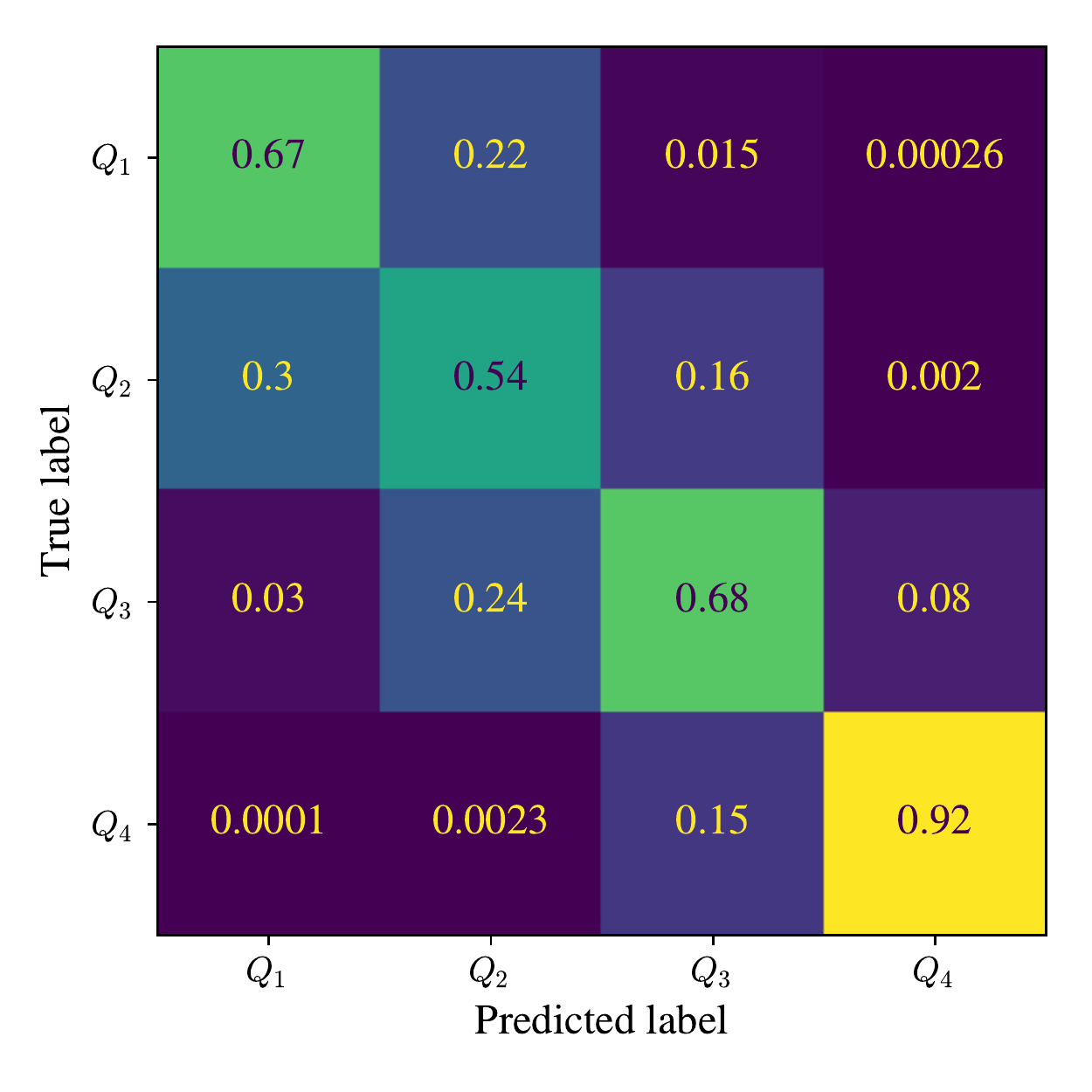}
	\caption{Confusion matrix with normalized columns for the best single stencil size model 
	with $s = 15$. Values on the diagonal are the correctly identified classes.}
	\label{fig:15ConfusionMatrix}
\end{figure}

Finally, the confusion matrix for the best model for $s = 15$ is shown in 
Fig.~\ref{fig:15ConfusionMatrix}. The results are very good and could be used 
in practice 
with 97\% of nodes classified as $Q_1$ below the median stencil error of 
$\epsilon = 3.9 \cdot 10^{-2}$ and 
practically 100\% of those classified as $Q_4$ above it.

\section{Future work}
\label{ch:future}

This is still a relatively untouched problem, so there is a plethora of 
future work remaining. We need to incorporate the machine learning based stencil evaluation into a node positioning algorithm and verify that the model can be used to improve the solution procedure and results on a benchmark PDE case. Additionally, steps should be taken to improve the model itself.

The first step would be to improve upon this work and achieve higher classification accuracies by utilising newer and more advanced models~\cite{guo2020pointCloudSurvey}, as the current results leave much to be desired if we were to use this classifier in practical stencil construction. Furthermore, for direct comparison between differently sized stencils, it would be beneficial if regression replaced classification, as we initially set out to do when starting this work, but neither attempts with multi layer perceptrons nor regressive modification for pointNet returned satisfactory results. Further work is required to determine whether regression would be feasible with the mentioned algorithms or if an alternative approach is required.

The second and probably larger task is to depart from deep learning towards 
something more explainable that could help us with deeper understanding 
of the many interconnected factors that contribute to a good stencil. This 
approach would require some work with feature engineering, but we believe 
that the benefits would be threefold: faster algorithm, better 
explainability and higher accuracy.

\section{Conclusions}
\label{ch:conclusions}
We have presented a methodology for computational generation of labelled 
stencil datasets that can be used for further data mining endeavours in stencil quality estimation. We have used the pointNet deep learning network to classify the quality of stencils, showing that it performs satisfactory as a proof of concept, but further work would be required before such classifier could be used to build new stencils. The most encouraging aspect of this approach is that the pointNet architecture allows for input padding without a negative 
impact on the results and thus allows us to use the same network for 
multiple stencil sizes with worse, but not drastically so, results. This 
could prove to be very beneficial as it avoids the necessity of having a 
dedicated network for each stencil size and could allow for cross-size 
optimisation.

\balance
\bibliographystyle{IEEEtran}
\bibliography{IEEEabrv, stencilML}

\end{document}